\documentclass{article}

\usepackage{arxiv}

\usepackage[utf8]{inputenc} 
\usepackage[T1]{fontenc}    
\usepackage{hyperref}       
\usepackage{url}            
\usepackage{booktabs}       
\usepackage{amsfonts}       
\usepackage{nicefrac}       
\usepackage{microtype}      
\usepackage{lipsum}

\RequirePackage{booktabs}

\usepackage{hyperref}
\usepackage{url}
\usepackage{subcaption}
\usepackage{graphicx} 
\usepackage{amsmath, bm}
\usepackage[ruled]{algorithm2e}
\usepackage{ amssymb }
\usepackage{bbm}
\usepackage{lipsum}  
\usepackage{multirow}

\usepackage{amssymb}
\usepackage{pifont}
\newcommand{\cmark}{\ding{51}}%
\newcommand{\xmark}{\ding{55}}%

\title{CUED\_Speech at TREC 2020 \\ Podcast Summarisation Track}

\author{Potsawee Manakul \\
  Engineering Department  \\
  University of Cambridge \\
  \texttt{pm574@cam.ac.uk} \\\And
  Mark Gales \\
  Engineering Department  \\
  University of Cambridge \\
  \texttt{mjfg@eng.cam.ac.uk} \\
}

\begin{document}

\maketitle

\begin{abstract}
In this paper, we describe our approach for the Podcast Summarisation challenge in TREC 2020. Given a podcast episode with its transcription, the goal is to generate a summary that captures the most important information in the content. Our approach consists of two steps: (1) Filtering redundant or less informative sentences in the transcription using the attention of a hierarchical model; (2) Applying a state-of-the-art text summarisation system (BART) fine-tuned on the Podcast data using a sequence-level reward function. Furthermore, we perform ensembles of three and nine models for our submission runs. We also fine-tune the BART model on the Podcast data as our baseline. The human evaluation by NIST\footnote{The National Institute of Standards and Technology (NIST)} shows that our best submission achieves 1.777 in the EGFB scale, while the score of creator-provided description is 1.291. Our system won the Spotify Podcast Summarisation Challenge in the TREC2020 Podcast Track in both human and automatic evaluation.   

\end{abstract}


\section{Introduction}
This paper describes our submissions to the TREC 2020 Podcast Track shared tasks \cite{trec2020podcastnotebook}. There are two tasks in the podcast track: ad-hoc segment retrieval (search), and summarisation. Both tasks use the Spotify Podcast dataset\footnote{\url{https://podcastsdataset.byspotify.com/}}. This work focuses on the summarisation task. The summarisation task involves generating a short text summary containing the most salient information for a given podcast episode with its audio and automatic transcription.

There are two main types of summarisation: \textit{extractive} methods which select and reorder words or sentences in the source; and \textit{abstractive} methods which can generate words and phrases that do not appear in the source. Due to the complex nature of abstractive methods, previous work on spoken document summarisation applied traditional machine learning or extractive methods \cite{svm_sum2013, towards_speech_sum, shang-etal-2018-unsupervised}. Recently, deep learning methods have shown much success and become the standard approach for various natural language processing (NLP) tasks, including abstractive text summarisation \cite{Goodfellow-et-al-2016}. In general, deep learning solves text summarisation by applying the encoder-decoder architecture \cite{sutskever2014sequence} based on recurrent neural networks (e.g. LSTM, GRU) \cite{nallapati-etal-2016-abstractive, see-etal-2017-get}, or self-attention networks \cite{liu-lapata-2019-text}. More recently, pre-training large transformer-based language model before fine-tuning the model on the target task has made a considerable improvement and achieved state-of-to-art results on various NLP tasks \cite{bert2018}. Based on this success, we propose that our simplest approach is to fine-tuning state-of-the-art abstractive text summarisation such as BART \cite{lewis-etal-2020-bart} on the podcast data. 

The limitation of pre-trained models is that they are trained on written text corpora, which generally have different characteristics from transcriptions of spoken documents. For instance, a transcription of 30-minute audio contains about 5,000 words\footnote{In this report, we will use the terms \textit{word} and \textit{token} interchangeably}, which is much longer than typical written text documents that are used during pre-training. As a result, commonly used pre-trained models such as BERT and BART can only take up to 512 or 1024 token due to their fixed maximum positional embedding. Another characteristic of spoken documents is that they contain redundancies in speech (e.g. false start, repetition) and some of the spoken utterances do not convey much information. We consider these issues as the long input sequence problem, and we will discuss it as well as proposing methods to handle in more details in Section \ref{section:bart_exp} and Section \ref{section:sentence_filtering}.

Furthermore, for this challenge, we explore the use of sequence-level training objectives described in Section \ref{section:sequence_level_training}, and ensemble of models described in Section \ref{section:ensemble}. In Section \ref{section:nist_eval}, we present and discuss the official evaluation of our submissions. Lastly, in Section \ref{section:conclusion}, we conclude our experiments as well as suggesting possible future work.

\section{Data}
\label{section:data}
More details about the Podcast dataset can be found in \cite{trec2020podcastnotebook}. This section presents the statistics of the podcast data that is useful in our design decisions. The dataset contains around 105,360 episodes from different podcast shows on Spotify. Google Cloud Platform's Cloud Speech-to-text API (GCP-ASR) was used to generate transcriptions. In this work, we do not make use of the audio data. We treat creator-provided descriptions as the summaries, so it should be noted that the quality of our target summaries varies from Bad (B) to Excellent (E). Some statistics are (mean $\pm$ standard deviation):
\begin{itemize}
    \item The number of words in a transcription: 5,727 $\pm$ 4,153
    \item The number of words in a summary: 61.1 $\pm$ 63.2
\end{itemize}
In our \textit{development} stage, we use the \textit{brass} subset of all the data (see \cite{trec2020podcastnotebook} for more details about processing steps done to obtain the brass set), resulting in 66,242 episodes in total. We further filtered out episodes with descriptions shorter than 5 tokens, and we process creator-provided descriptions by removing URL links and \verb|@name|. Then, we split the data into train/dev sets of 60,415/2,189 episodes. In addition, 150 episodes were selected with 6 set of summaries for each episode (900 document-summary-grade triplets), and they were graded are on the Bad/Fair/Good/Excellent scale (0-3). We use this subset to train our grader.

In the \textit{evaluation} stage, the evaluation (test) set consists of 1,027 episodes, and we will test some models and ensembles of models that are trained on the train set of the development stage on this evaluation dataset.

\section{Methodology}
In this section, we describe our methods using the following notation: $\boldsymbol{\theta}$ is the summarisation model parameter, $\{ (\mathbf{x}^{(1)}, \mathbf{y}^{(1)}), (\mathbf{x}^{(2)}, \mathbf{y}^{(2)}),..., (\mathbf{x}^{(J)}, \mathbf{y}^{(J)}) \}$ is $J$ transcription-summary pairs. For simplicity of notation, we will express losses and probability distributions for one pair of data $(\mathbf{x}, \mathbf{y})$.

\subsection{Baseline: Fine-tuning BART Model}
\label{section:bart_exp}
The BART model was proposed in \cite{lewis-etal-2020-bart}. It has a standard sequence-to-sequence architecture with a bidirectional encoder (similar to BERT \cite{bert2018}) and a causal decoder (i.e. left-to-right) decoder (similar to GPT-2 \cite{radford2019language}). Because of the bidirectional nature of the encoder, and the causal nature of the decoder, BART is suitable for conditional text generation tasks. BART is pre-trained using an unsupervised criterion by randomly shuffling the order in the original text as well as filling masked tokens. In this work, we use \verb|bart-large|\footnote{\url{https://huggingface.co/}} consisting of token embedding (shared among encoder, decoder, and language model head), positional embedding, 12-layer encoder, 12-layer decoder, hidden size of 1024, 16 heads (406M parameters in total). Each token to the encoder or decoder is embedded by its token\_id and position. The BART model is firstly tuned on a news summarisation corpus such as CNN/DailyMail \cite{nallapati-etal-2016-abstractive} or XSum \cite{narayan-etal-2018-dont}, before being fine-tuned on the podcast data. Training (or fine-tuning) is done by minimising the negative log-likelihood of target sequence as follows: 
    \begin{equation}
        \mathcal{L}_{\text{ml}} = -  \log P\left(  \mathbf{y} | \mathbf{x}  ;{\bm\theta}\right) = -  \sum_t \log P\left(  y_t | \mathbf{y}_{1:t-1}, \mathbf{x}  ;{\bm\theta}\right)
    \end{equation}

The positional embedding uses the absolute position, limiting the length of the input sequence to 1,024 tokens. However, Section \ref{section:data} shows that the average number of tokens in a podcast transcription is 5,727; thus, a considerable amount of information is lost when truncating the transcription to be within 1,024 tokens. Thus, firstly we propose to expand the positional embedding of the BART model to accommodate sequences longer than 1,024, and we compare it to the vanilla BART model.

\subsubsection*{Positional embedding expansion}
We create a new positional embedding matrix of a size larger than 1,024 tokens; we copy the learned positional embedding weight of the 1st to the 1,024th positions and randomly initialised those beyond 1,024th position. Then, we can train the BART model end-to-end on the podcast data without or with less aggressive truncation.

In this experiment, due to the limited amount of GPU memory (11 GB), we expand the maximum number of tokens to 3,954; we freeze parts of the BART model and fine-tune the last layers of the encoder and decoder as well as the language model head (81M parameters in total). In the first two blocks of Table \ref{tab:filtering_length_limit}, we show that the non-expanded BART performs better than the expanded BART, and the reasons are likely due to that: (1) since BART uses a learned positional embedding, instead of sinusoidal function in the original Transformer, simply expanding it does not yield a good performance; (2) learning very long sequences could be inefficient. Lastly, we show in the final block of Table \ref{tab:filtering_length_limit} that training all parameters of BART but keeping the maximum positional embedding at 1,024 tokens yields the highest ROUGE scores. Therefore, we conclude that expanding positional embedding is \textit{inefficient}, and we will use the vanilla BART model. 

\begin{table}[h!]
  \centering
  \begin{tabular}{ccc|ccc}
    \toprule
   \multicolumn{3}{c}{\textbf{BART Configuration}}  &\multicolumn{3}{c}{\textbf{ROUGE F$_1$}} \\
    \textbf{MaxLen} &\textbf{Part frozen} &\textbf{\#Param} &\textbf{R-1} &\textbf{R-2} &\textbf{R-L}  \\
    \midrule

    3954  &\cmark  &81M &26.72    &7.70  &22.87 \\
    1024  &\cmark  &81M   &27.61  &8.88  &23.99 \\
    1024  &\xmark  &406M  &29.09  &9.92  &25.37   \\
    \bottomrule
  \end{tabular}
  \caption{Variants of BART Model Comparison. (1) Expanded BART: BART with maximum positional embedding of 3,954, (2) Non-expanded BART with original maximum positional embedding of 1,024 but parts of the model is frozen the same way as Expanded-BART, (3) Vanilla BART. All systems are trained on the train split of the brass set, and evaluated on the dev split of the brass set.}
  \label{tab:filtering_length_limit}
\end{table}

\subsection{Sentence Filtering}
\label{section:sentence_filtering}
The memory and time required for full self-attention models grow quadratically with the input sequence length $O(n^2)$, preventing us from fine-tuning the model end-to-end on the entire input transcriptions. Also, in Section \ref{section:bart_exp}, we show that expanding positional embedding does not improve the performance. Thus, we investigate alternative methods by \textit{filtering} out redundant or less informative sentences in the input transcriptions.
\begin{enumerate}
    \item \textit{Random}: Randomly select input sentences such that the length does not exceed the maximum length. Note that we keep the same sentence order.
    \item \textit{Truncation}: The simplest method is to select the first $n$ tokens in the input transcription.
    \item \textit{TextRank} \cite{mihalcea2004textrank}: The algorithm ranks sentences based on their similarity to other sentences. In this work, we represent each sentence by averaging \verb|word2vec| embedding. We select top sentences such that in total there are less than $n$ tokens, and we keep the same sentence order.
    \item \textit{Hierarchical Attention (HIER)}: We train the hierarchical encoder decoder model used in \cite{manakul2020abstractive} on podcast data without truncation. At test time, we sum the sentence-level attention score over all time instances to obtain a measure of the importance of sentence $i$:
        \begin{equation}
            v_i = \frac{1}{T}\sum_{t=1}^T \alpha^{\tt s}_{t,i}
            \label{eq:hier_model_attn_score}
        \end{equation} 
    where $\alpha^{\tt s}_{t,i}$ is the sentence-level attention score of the hierarchical model, and $t$ is the decoder time instance. Note that if the target sequence is available (e.g. in training data), we can use teacher forcing decoding to obtain the importance score. However, if the target sequence is not available, we have to rely on a decoding method such as beam search.
\end{enumerate}
Sentence filtering can be applied on the input transcriptions at the training time and/or the test time, so we will look at the following scenarios: 
\begin{itemize}
    \item \textbf{Test-time only: Comparing different filtering methods}\\
    During training, the podcast transcription is truncated to be within the maximum length of 1,024 tokens. We compare the filtering methods described above at test time: Random, Truncate, TextRank, and HIER. The hierarchical model (HIER) is trained on CNN/DailyMail and fine-tuned on the podcast data. Results in Table \ref{tab:filtering_methods} show that using the attention of the hierarchical model yields the highest ROUGE scores; truncation is better than random selection; however, the TextRank algorithm yields the worst results. 
    
\begin{table}[h!]
  \centering
  \begin{tabular}{l|ccc}
    \toprule
     \textbf{Filtering Method} &\multicolumn{3}{c}{\textbf{ROUGE F$_1$}} \\
    \textbf{Test-time}  &\textbf{R-1} &\textbf{R-2} &\textbf{R-L}  \\
    \midrule
    Random              &25.65  &6.89  &22.10 \\    
    Truncate            &29.09  &9.92  &25.37 \\
    TextRank            &24.97  &6.40  &21.47 \\
    HIER                &\textbf{29.37}  &\textbf{10.02}  &\textbf{25.51} \\
    \bottomrule
  \end{tabular}
  \caption{Comparison of filtering methods at test time. We use the vanilla BART system with maximum positional embedding of 1,024. At training time, the input transcription is truncated and all model parameters are fine-tuned, but at test time the input transcription is filtered.}
  \label{tab:filtering_methods}
\end{table}

    \item \textbf{Training-time and Test-time filtering}\\
Here, we investigate another option which is to filter the data at the training time in addition to the test time. We select the hierarchical model (HIER) as our filtering method, and at training time we could either use filtered data to train BART from scratch (Train) or fine-tune it on BART trained on truncated data (Fine-tune). Table \ref{tab:filtering_train_and_test_time} shows that the best summarisation performance is achieved when we perform filtering by both training and test times. Another observation is that with the filtered data training a model from scratch is better than continue the training from a model trained on truncated data, and this could be due to a local optimum in training.

\begin{table}[h!]
  \centering
  \begin{tabular}{ll|ccc}
    \toprule
    \multicolumn{2}{c}{\textbf{Filtering Method}} &\multicolumn{3}{c}{\textbf{ROUGE F$_1$}} \\
   \textbf{Training-time} &\textbf{Test-time} &\textbf{R-1} &\textbf{R-2} &\textbf{R-L}  \\
    \midrule
    Truncate          &Truncate  &29.09  &9.92  &25.37   \\
    Truncate          &HIER    &29.37  &10.02 &25.51   \\
    HIER (Fine-tune)  &HIER    &29.38  &10.04 &25.69   \\
    HIER (Train)      &HIER    &\textbf{29.74}  &\textbf{10.25} &\textbf{25.71}   \\
    \bottomrule
  \end{tabular}
  \caption{Effectiveness of sentence filtering at training time and test time.}
  \label{tab:filtering_train_and_test_time}
\end{table}
\end{itemize}

We have shown the effectiveness of sentence filtering at both training time and test time using the hierarchical model. Note that sentence filtering can be thought of as a less aggressive extractive summarisation method. Typically, extractive summarisation aims to select most $N$ (e.g. $N=3$) salient sentences by creating pseudo labels \cite{liu-lapata-2019-text}. From now on, we will make use of the hierarchical model to filter input transcriptions at both training and test time.

\subsection{Sequence-level Loss Training}
\label{section:sequence_level_training}
During training, we maximise the likelihood at the token level, while during test time, we use automatic metrics such as ROUGE or we evaluate manually. Both automatic and manual evaluations at test time operate at the sequence level. Thus, optimising at the token level only as done in maximum likelihood training is not expected to yield the best result. We follow \cite{paulus2017deep}, which uses reinforcement learning inspired loss:
\begin{equation}
    \mathcal{L}_{rl} = \left( \text{Reward}(\tilde{\mathbf{y}}) -  \text{Reward}(\hat{\mathbf{y}})   \right) \sum_t \log P(\hat{y}_t | \hat{\mathbf{y}}_{1:t-1},\mathbf{x} ; \boldsymbol{\theta})     
\end{equation}

where $\hat{\mathbf{y}}$ is the sequence obtained by sampling $\hat{y}_t \sim P(y| \hat{\mathbf{y}}_{1:t-1}, \mathbf{x} ; \boldsymbol{\theta})$ at each time step, and $\tilde{\mathbf{y}}$ is the sequence obtained by greedy search. In \cite{paulus2017deep}, training on the sequence-level loss directly is not stable, so we initialise the model using the model weights trained on maximum likelihood. We train the model on the sequence-level loss function in combination with  maximum likelihood loss:
\begin{equation}
    \mathcal{L} = \gamma\mathcal{L}_\text{rl} + (1-\gamma)\mathcal{L}_\text{ml}
\end{equation}

We experiment two types of the reward function to compute $\mathcal{L}_{rl}$:
\begin{enumerate}
    \item \textit{ROUGE-L}: we follow the ROUGE-L reward training as proposed in \cite{paulus2017deep}. The ROUGE score is computed against the target summary $\mathbf{y}$: $\text{Reward}(\tilde{\mathbf{y}}) = \text{ROUGE-L}(\tilde{\mathbf{y}}, \mathbf{y})$, and $\text{Reward}(\hat{\mathbf{y}}) = \text{ROUGE-L}(\hat{\mathbf{y}}, \mathbf{y})$.
    \item \textit{Automatic Grader}: we train a neural grader model $\boldsymbol{\phi}$ that predicts the grade given a pair of document and summary: $\text{Reward}(\tilde{\mathbf{y}}) = f_{\boldsymbol{\phi}}(\mathbf{x}, \tilde{\mathbf{y}})$, and $\text{Reward}(\hat{\mathbf{y}}) = f_{\boldsymbol{\phi}}(\mathbf{x},\hat{\mathbf{y}})$, where $f_{\boldsymbol{\phi}}(.)$ gives the grade predicted by model $\boldsymbol{\phi}$.
\end{enumerate}

The grader is a convolutional neural network (CNN) model that takes a similarity matrix between input transcription and summary as the input. Each cell $(i,j)$ in the similarity matrix is computed by cosine similarity between the vector representing sentence $i$ in the input, and the vector representing sentence $j$ in the summary. We use Sentence-BERT \cite{reimers-2019-sentence-bert} for the sentence representation. We can use multiple variants of Sentence-BERT models for $M$-dimensional similarity matrices. In this experiment, we try $M=1$ using \verb|bert-large-nli-mean-tokens|, and $M=10$ using nine additional variants. We train the grader model on the part of the podcast data that are annotated with grade information (900 examples from 150 episodes, each episode contains 6 sets of summaries).

\begin{figure}[h!]
    \centering
    \begin{subfigure}[b]{0.49\textwidth}
         \centering
\includegraphics[width=\linewidth,height=6cm,keepaspectratio]{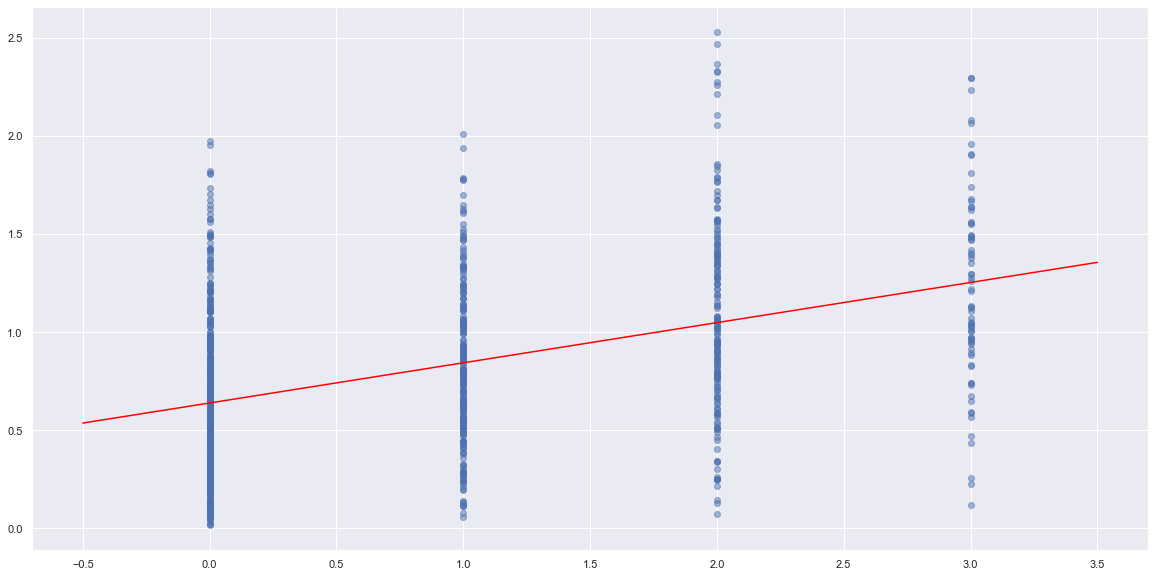}
  \caption{1-channel input, PCC=0.421}
    \end{subfigure}%
    \hfill
    \begin{subfigure}[b]{0.49\textwidth}
     \centering
      \includegraphics[width=\linewidth,height=6cm,keepaspectratio]{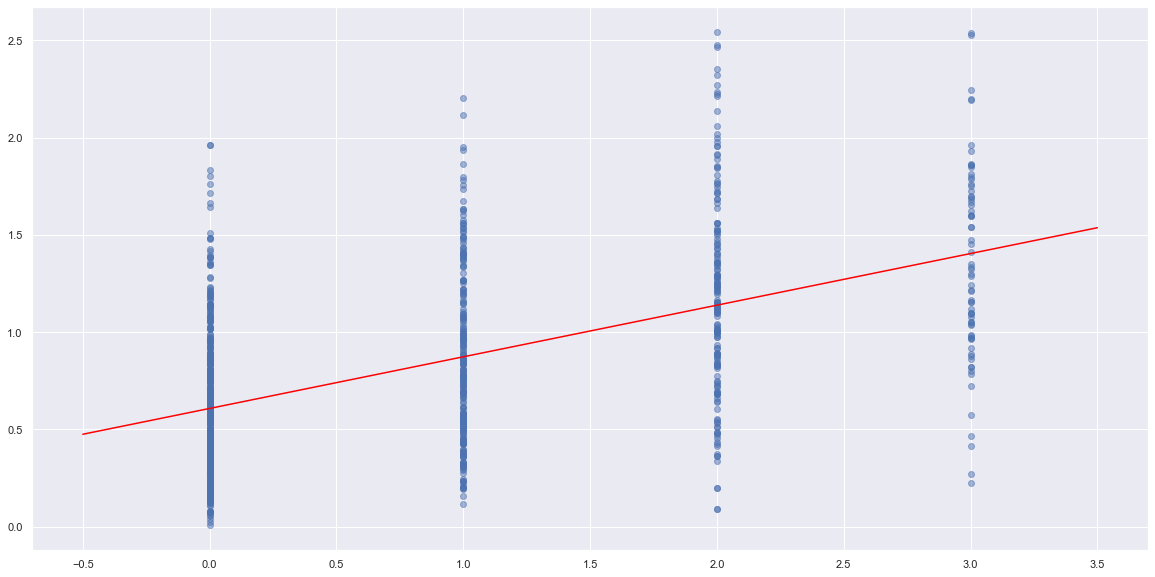}
  \caption{10-channel input, PCC=0.517}
    \end{subfigure}%
    \hfill
  \caption{9-fold cross validation experiment on the subset of podcast data that has grade information.}
  \label{fig:reg_exp}
\end{figure}

As illustrated in Figure \ref{fig:reg_exp}, we perform 9-fold cross-validation on the subset of podcast data that has grade information, and we achieve Pearson Correlation Coefficient (PCC) of 0.421 for 1-channel input, and that of 0.517 for 10-channel input. Subsequently, we train our CNN model on all the data (900 graded episodes) using 1-channel input due to limited computational resource when using the model for reward optimisation. 

In Table \ref{tab:rl_bart_training}, we show that using ROUGE-L as the reward is more effective, and there is an improvement over the baseline. When using our automatic grader; however, the summarisation model achieves higher ROUGE-2 and ROUGE-L scores than the baseline, but this model does not outperform the model trained using ROUGE-L. We believe that the outputs of the automatic-grader-reward model may be better than those of the ROUGE-L-reward model, but further human analysis is required to make a conclusion. Hence, we decided to use the ROUGE-L-reward model to our final ensemble, which is described in the next section.

\begin{table}[ht!]
  \centering
  \begin{tabular}{lc|ccc}
    \toprule
    &  &\multicolumn{3}{c}{\textbf{ROUGE F$_1$}} \\
   \textbf{System} &\textbf{Reward} &\textbf{R-1} &\textbf{R-2} &\textbf{R-L}  \\
    \midrule
    ML-trained         &-          &29.74 &10.25 &25.71  \\
    \midrule
    $\gamma=1.0$     &ROUGE-L    &\textbf{29.97} &10.32 &25.86  \\
    $\gamma=0.9$     &ROUGE-L    &29.92 &\textbf{10.42} &25.90  \\
    $\gamma=1.0$     &Grader     &29.70   &10.35    &\textbf{25.97}    \\
    $\gamma=0.9$     &Grader     &29.73   &10.20    &25.80    \\
    \bottomrule
  \end{tabular}
  \caption{The training data is filtered using HIER model, and the baseline is the model trained on filtered data using only $\mathcal{L}_\text{ml}$ criterion. Note that to fit the training on one 11GB GPU, the first three layers of the encoder and decoder are frozen.}
  \label{tab:rl_bart_training}
\end{table}

\subsection{Ensemble of models}
\label{section:ensemble} 
We obtain different set of model's weights from different data shuffles, and checkpoints. Given $M$ systems $\{\boldsymbol{\theta}^{1},...,\boldsymbol{\theta}^{M}\}$, we combine the predictive distribution on the input $\mathbf{x}$ at token-level as follows:
        \begin{equation}
            P(\mathbf{y}^* | \mathbf{x}) =  \prod_{t=1}^T P(y_t^* | \mathbf{y}^*_{1:t-1}, \mathbf{x}) =   \prod_{t=1}^T \left[ \frac{1}{M} \sum_{m=1}^M  P(y_t^* | \mathbf{y}^*_{1:t-1}, \mathbf{x} ; \boldsymbol{\theta}^{m}) \right]
            \label{eq:combine_token_level}
        \end{equation}
where $\mathbf{y}^*$ denotes the hypothesis sequence.

Having investigated various approaches on our own development set, in this part we evaluate their performance on the \textbf{evaluation} (test) set of 1,027 episodes. We also build ensembles of models by using different data shuffles, and parameters at different checkpoints. During the evaluation stage, we found that fining-tuning the BART summarisation model tuned on the XSum dataset yields a better result than fine-tuning the same model tuned on the CNN/DailyMail as shown in Table \ref{tab:main_table}.
\begin{table}[ht!]
  \centering
  \begin{tabular}{l|cc|ccc}
    \toprule
    &\multicolumn{2}{c}{\textbf{Filtering}} &\multicolumn{3}{c}{\textbf{ROUGE F$_1$}} \\
   \textbf{System}          &\textbf{Train}   &\textbf{Test}  &\textbf{R-1} &\textbf{R-2} &\textbf{R-L}($\dagger$)   \\
    \midrule
    Baseline*  &\xmark   &\xmark      &26.57    &9.14     &23.43 (18.44)  \\
    Baseline   &\xmark   &HIER        &26.96    &9.53     &23.70 (18.57)  \\
    Filtered   &HIER     &HIER        &26.96    &9.75     &23.71 (18.90)  \\

    \midrule
    Filtered(XSum)*  &HIER  &HIER     &27.10    &9.96     &23.92 (18.91)  \\
    Filtered(XSum)+$\mathcal{L}_{rl}$ &HIER    &HIER       &27.91     &10.25     &24.67 (19.48)  \\
    \midrule
    Ensemble(3shuffles)*      &HIER   &HIER      &28.56    &10.83     &25.23 (19.98) \\
    Ensemble(3checkpoints)   &HIER    &HIER      &28.12     &10.44     &24.87 (19.66) \\
   Ensemble(3shuffles$\times$3checkpoints)* &HIER  &HIER     &\textbf{28.57}    &\textbf{10.86}     &\textbf{25.28 (19.98)}  \\
    \bottomrule
  \end{tabular}
  \caption{ROUGE F$_1$ scores on the evaluation set (1,027 episodes). We use our processed creator-provided descriptions as described in Section \ref{section:data} as the reference for computing ROUGE scores. We split the summaries into sentences to compute longest n-grams only within sentences; hence the first ROUGE-L numbers, and the ROUGE-L numbers in ($\dagger$) are obtained when the summary is treated as one sequence. We compare our ROUGE-L scores against NIST-evaluated scores in Table \ref{tab:nist_rouge}. *The summaries of these systems were submitted.}
  \label{tab:main_table}
\end{table}

\section{NIST Evaluation Results}
\label{section:nist_eval}
We submitted four set of summaries from the following systems:
\begin{enumerate}
    \item {CUED-baseline}: run\_id=\verb|cued_speechUniv3| - BART model fine-tuned on truncated podcast data.
    \item {CUED-filtered}: run\_id=\verb|cued_speechUniv4| - BART model fine-tuned on the transcriptions filtered using the hierarchical model.
    \item {CUED-ensemble3}: run\_id=\verb|cued_speechUniv2| - Ensemble of 3 BART models (3 random seeds $\times$ 1 checkpoints), each trained on filtered transcription (using hierarchical model) data + $\mathcal{L}_{rl}$ criterion
    \item {CUED-ensemble9}: run\_id=\verb|cued_speechUniv1| - Ensemble of 9 BART models (3 random seeds $\times$ 3 checkpoints), each trained on filtered transcription (using hierarchical model) data + $\mathcal{L}_{rl}$ criterion
\end{enumerate}
We are also provided with results of two baselines from Spotify:
\begin{enumerate}
    \item SPFT-desc: creator-provided descriptions as summaries
    \item SPFT-filt: filtered creator-provided descriptions as summaries
\end{enumerate}

179 episodes were selected randomly, and NIST annotators evaluated them on the Bad/Fair/Good/Excellent scale, and answered eight yes/no question (Q1-Q8 in Table \ref{tab:nist_eval}. For example, Q1 is "Does the summary include names of the main people (hosts, guests, characters)
involved or mentioned in the podcast?".

\begin{figure}[ht!]
    \centering
    \begin{subfigure}[b]{0.50\textwidth}
         \centering
      \includegraphics[width=\linewidth,height=6cm,keepaspectratio]{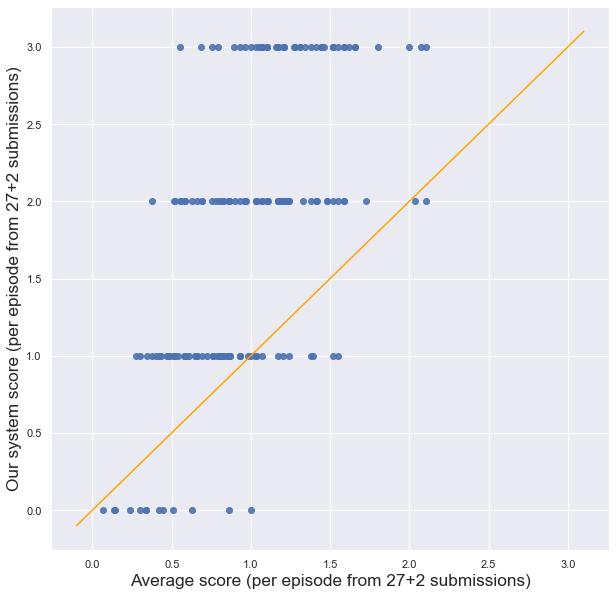}
      \caption{CUED-ensemble3 (84.36\%)}
    \end{subfigure}%
    \hfill
    \begin{subfigure}[b]{0.50\textwidth}
     \centering
      \includegraphics[width=\linewidth,height=6cm,keepaspectratio]{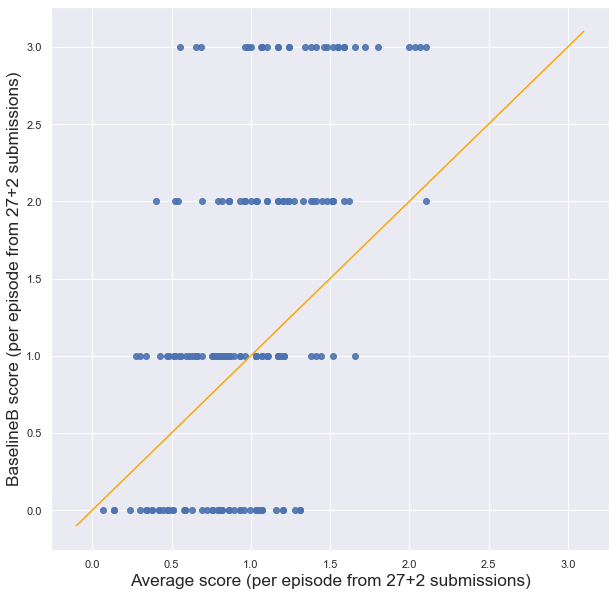}
      \caption{SPTF-filt, (60.34\%)}
    \end{subfigure}%
    \hfill
  \caption{Per summary analysis of our submissions. There are in total 27+2 sets of summaries. X-axis is the average score of all submissions, and Y-axis is the system score. The number in brackets is the number of times that the system achieves a higher score than the average score.}
  \label{fig:per_summary_podcast}
\end{figure}

Table 6 shows that our CUED-ensemble3 submission receives the highest human rating at 1.777 in average, compared to our CUED-baseline at 1.564 and SPFT-desc at 1.291. When using an automatic evaluation, CUED-ensemble3 also achieves the highest score shown in Table \ref{tab:nist_rouge} (note that our ROUGE scores, computed using processed episode descriptions, suggest the performance of CUED-ensemble3 and CUED-ensemble9 are similar). Furthermore, the per episode analysis (in Figure \ref{fig:per_summary_podcast}) shows that CUED-ensemble3 performs better than an \textit{average} model\footnote{We obtain the score of an average model by computing the mean score of each episode} on 84.36\% of the judged episodes, compared to 79.33\% for CUED-baseline, and 60.34\% for SPTF-filt.

\begin{table}[!ht]
  \centering
  \begin{tabular}{l|cccc|cccccccc}
    \toprule
    \textbf{System}            &\textbf{Avg}    &\textbf{\%E}  &\textbf{\%EG}   &\textbf{\%EGF}    &\textbf{Q1}  &\textbf{Q2} &\textbf{Q3}  &\textbf{Q4}  &\textbf{Q5}  &\textbf{Q6}($\downarrow$)  &\textbf{Q7}  &\textbf{Q8} \\
    \midrule
    SPTF-desc         &1.291      &15.6   &39.7  &73.7      &55.9	&34.1	&72.1	&55.3	&63.7	&3.4	&77.7	&52.0      \\
    SPTF-filt         &1.307      &18.4   &39.7  &72.6      &58.1	&38.0	&70.4	&53.1	&60.3	&2.8	&78.2	&60.9      \\
    \midrule
    CUED-baseline     &1.564      &21.8   &50.3  &84.4      &\textbf{63.7}	&\textbf{44.1}	&82.7	&61.5	&74.9	&\textbf{6.7}	&88.3	&70.4      \\
    CUED-filtered     &1.687      &25.7   &56.4  &86.6      &62.0	&43.0	&87.2	&\textbf{63.1}	&79.3	&7.8	&\textbf{88.8}	&73.7      \\
    CUED-ensemble3    &\textbf{1.777}      &26.3   &\textbf{58.7}  &\textbf{92.7}      &\textbf{63.7}	&39.1	&\textbf{89.9}	&\textbf{63.1}	&\textbf{80.4}	&7.8	&88.3	&\textbf{77.1}      \\
    CUED-ensemble9    &1.704      &\textbf{27.4}   &58.1  &84.9      &62.0	&40.8	&86.0	&60.3	&\textbf{80.4}	&10.1	&86.6	&76.5      \\
    \bottomrule
  \end{tabular}
  \caption{Human Evaluation Results. Avg = average of E=3, G=2, F=1, B=0. \%E is the percentage of episodes graded excellent, \%EG is the percentage of episodes graded excellent or good, \%EGF is the percentage of episodes graded excellent, good, or fair. Q1-Q8 are the percentage of \textit{yes} answers.}
  \label{tab:nist_eval}
\end{table}

\begin{table}[ht!]
  \centering
  \begin{tabular}{l|cc}
    \toprule
     &\multicolumn{2}{c}{\textbf{ROUGE F$_1$}} \\
    \textbf{System}            &\textbf{R-L(NIST)}  &\textbf{R-L(ours)}* \\
    \midrule
    CUED-baseline    &18.120            &18.44   \\
    CUED-filtered    &18.486            &18.91   \\
    CUED-ensemble3   &\textbf{19.670}   &19.98   \\
    CUED-ensemble9   &19.661            &19.98   \\
    \bottomrule
  \end{tabular}
  \caption{Automatic Evaluation Results using the ROUGE-L (F$_1$) metric. *Our reference for computing ROUGE-L is based on our processed version of episode descriptions (after remove URL links, etc), and we believe that the processed version should better reflect what the models are trained as it is less noisy.}
  \label{tab:nist_rouge}
\end{table}

\newpage
\section{Conclusion and Future Work}
\label{section:conclusion}
Our best system consists of (1) using the hierarchical model to filter transcriptions at both training time and test time, (2) fine-tuning the BART model initialised on XSum without expanding its positional embedding, (3) optimising the sequence-level loss $\mathcal{L}_{rl}$ after optimising the token-level maximum likelihood, (4) ensemble of models. Both automatic and human evaluations show that this approach outperforms our baseline. Our system-generated summaries obtain higher human ratings than the target summaries (episode descriptions), suggesting training the model using episode descriptions as target summaries can be improved. For instance, an automatic grader can be used to select if an episode description is appropriate as the target or not, allowing one to use semi-supervised learning approaches. Furthermore, the long sequence problem can be better addressed by either improving sentence filtering or using other neural architectures without full self-attention mechanism. Lastly, there is additional information in audio that are not utilised in this work. 
\section*{Acknowledgments}
This paper reports on research supported by ALTA institute, Cambridge Assessment English, University of Cambridge. Thanks to Yiting Lu for evaluating our system-generated summaries before submission. 

\bibliographystyle{unsrt}  
\bibliography{references}  

\begin{thebibliography}{10}

\bibitem{trec2020podcastnotebook}
Ann Clifton, Aasish Pappu, Sravana Reddy, Yongze Yu, Jussi Karlgren, Ben
  Carterette, Rosie Jones, Maria Eskevich, and Gareth Jones.
\newblock {O}verview of the {TREC} 2020 {P}odcasts {T}rack.
\newblock In {\em The 29th {T}ext {R}etrieval {C}onference (TREC 2020)
  notebook}. NIST, 2020.

\bibitem{svm_sum2013}
Sz-Rung Shiang, H.-Y Lee, and L.-S Lee.
\newblock Supervised spoken document summarization based on structured support
  vector machine with utterance clusters as hidden variables.
\newblock {\em Proceedings of the Annual Conference of the International Speech
  Communication Association, INTERSPEECH}, pages 2728--2732, 2013.

\bibitem{towards_speech_sum}
F.~{Liu} and Y.~{Liu}.
\newblock Towards abstractive speech summarization: Exploring unsupervised and
  supervised approaches for spoken utterance compression.
\newblock {\em IEEE Transactions on Audio, Speech, and Language Processing},
  21(7):1469--1480, 2013.

\bibitem{shang-etal-2018-unsupervised}
Guokan Shang, Wensi Ding, Zekun Zhang, Antoine Tixier, Polykarpos Meladianos,
  Michalis Vazirgiannis, and Jean-Pierre Lorr{\'e}.
\newblock Unsupervised abstractive meeting summarization with multi-sentence
  compression and budgeted submodular maximization.
\newblock In {\em Proceedings of the 56th Annual Meeting of the Association for
  Computational Linguistics}, 2018.

\bibitem{Goodfellow-et-al-2016}
Ian Goodfellow, Yoshua Bengio, and Aaron Courville.
\newblock {\em Deep Learning}.
\newblock MIT Press, 2016.
\newblock \url{http://www.deeplearningbook.org}.

\bibitem{sutskever2014sequence}
Ilya Sutskever, Oriol Vinyals, and Quoc~V Le.
\newblock Sequence to sequence learning with neural networks.
\newblock In {\em Advances in neural information processing systems}, pages
  3104--3112, 2014.

\bibitem{nallapati-etal-2016-abstractive}
Ramesh Nallapati, Bowen Zhou, Cicero dos Santos, {\c{C}}a{\u{g}}lar
  GuÌ‡l{\c{c}}ehre, and Bing Xiang.
\newblock Abstractive text summarization using sequence-to-sequence {RNN}s and
  beyond.
\newblock In {\em Proceedings of The 20th {SIGNLL} Conference on Computational
  Natural Language Learning}. Association for Computational Linguistics, August
  2016.

\bibitem{see-etal-2017-get}
Abigail See, Peter~J. Liu, and Christopher~D. Manning.
\newblock Get to the point: Summarization with pointer-generator networks.
\newblock In {\em Proceedings of the 55th Annual Meeting of the Association for
  Computational Linguistics}, July 2017.

\bibitem{liu-lapata-2019-text}
Yang Liu and Mirella Lapata.
\newblock Text summarization with pretrained encoders.
\newblock In {\em Proceedings of the 2019 Conference on Empirical Methods in
  Natural Language Processing and the 9th International Joint Conference on
  Natural Language Processing (EMNLP-IJCNLP)}, 2019.

\bibitem{bert2018}
Jacob Devlin, Ming{-}Wei Chang, Kenton Lee, and Kristina Toutanova.
\newblock {BERT:} pre-training of deep bidirectional transformers for language
  understanding.
\newblock {\em CoRR}, abs/1810.04805, 2018.

\bibitem{lewis-etal-2020-bart}
Mike Lewis, Yinhan Liu, Naman Goyal, Marjan Ghazvininejad, Abdelrahman Mohamed,
  Omer Levy, Veselin Stoyanov, and Luke Zettlemoyer.
\newblock {BART}: Denoising sequence-to-sequence pre-training for natural
  language generation, translation, and comprehension.
\newblock In {\em Proceedings of the 58th Annual Meeting of the Association for
  Computational Linguistics}, pages 7871--7880, Online, July 2020. Association
  for Computational Linguistics.

\bibitem{radford2019language}
Alec Radford, Jeffrey Wu, Rewon Child, David Luan, Dario Amodei, and Ilya
  Sutskever.
\newblock Language models are unsupervised multitask learners.
\newblock {\em OpenAI Blog}, 1(8):9, 2019.

\bibitem{narayan-etal-2018-dont}
Shashi Narayan, Shay~B. Cohen, and Mirella Lapata.
\newblock Don{'}t give me the details, just the summary! topic-aware
  convolutional neural networks for extreme summarization.
\newblock In {\em Proceedings of the 2018 Conference on Empirical Methods in
  Natural Language Processing}, pages 1797--1807, Brussels, Belgium,
  October-November 2018. Association for Computational Linguistics.

\bibitem{mihalcea2004textrank}
Rada Mihalcea and Paul Tarau.
\newblock Textrank: Bringing order into text.
\newblock In {\em Proceedings of the 2004 conference on empirical methods in
  natural language processing}, pages 404--411, 2004.

\bibitem{manakul2020abstractive}
Potsawee Manakul, Mark~J.F. Gales, and Linlin Wang.
\newblock {Abstractive Spoken Document Summarization Using Hierarchical Model
  with Multi-Stage Attention Diversity Optimization}.
\newblock In {\em Proc. Interspeech 2020}, pages 4248--4252, 2020.

\bibitem{paulus2017deep}
Romain Paulus, Caiming Xiong, and Richard Socher.
\newblock A deep reinforced model for abstractive summarization.
\newblock {\em arXiv preprint arXiv:1705.04304}, 2017.

\bibitem{reimers-2019-sentence-bert}
Nils Reimers and Iryna Gurevych.
\newblock Sentence-bert: Sentence embeddings using siamese bert-networks.
\newblock In {\em Proceedings of the 2019 Conference on Empirical Methods in
  Natural Language Processing}. Association for Computational Linguistics, 11
  2019.

\end{thebibliography}
\end{document}